\documentclass[cmc,article,accept,moreauthors,pdftex]{Definitions/tsp}
\usepackage[normalem]{ulem} 
\usepackage{amsfonts} 
\usepackage{mathcomp} 
\usepackage{CJKutf8} 
\usepackage{pifont} 
\usepackage{bm} 
\usepackage{bbm} 
\graphicspath{{./Definitions/}} 

\makeatletter
\def\T@n@@nc@d@ngM@cr@M@d{}
\def\LY@n@@nc@d@ngM@cr@M@d{}
\makeatother

\let\orignewcommand\newcommand  
\let\newcommand\providecommand  
\usepackage{verse}
\let\newcommand\orignewcommand  

\newsavebox\foobox

\renewcommand{\dagger}{\mathchar"2279}
\renewcommand{\ddagger}{\mathchar"227A}

\newcommand{\mmathit}[1]{
  \ifthenelse{\equal{#1}{\ln}}{\mathit{ln}}{
    \ifthenelse{\equal{#1}{\max}}{\mathit{max}}{\mathit{#1}}
  }
}
\makeatother
\robustify{\footnote}

\DeclareUnicodeCharacter{1E45}{\.{n}}
\DeclareUnicodeCharacter{1E41}{\.{m}}
\DeclareUnicodeCharacter{2003}{\quad}
\DeclareUnicodeCharacter{2009}{\thinspace}
\DeclareUnicodeCharacter{2002}{\enspace{}}
\DeclareUnicodeCharacter{2005}{\thinspace}
\DeclareUnicodeCharacter{0263}{\textipa{G}}
\DeclareUnicodeCharacter{A0}{~}
\DeclareUnicodeCharacter{2460}{\textcircled{\scriptsize{1}}}
\DeclareUnicodeCharacter{2461}{\textcircled{\scriptsize{2}}}
\DeclareUnicodeCharacter{2462}{\textcircled{\scriptsize{3}}}
\DeclareUnicodeCharacter{2463}{\textcircled{\scriptsize{4}}}
\DeclareUnicodeCharacter{2464}{\textcircled{\scriptsize{5}}}
\DeclareUnicodeCharacter{2465}{\textcircled{\scriptsize{6}}}
\DeclareUnicodeCharacter{2466}{\textcircled{\scriptsize{7}}}
\DeclareUnicodeCharacter{2467}{\textcircled{\scriptsize{8}}}
\DeclareUnicodeCharacter{2468}{\textcircled{\scriptsize{9}}}
\DeclareUnicodeCharacter{2070}{\textsuperscript{0}}
\DeclareUnicodeCharacter{2074}{\textsuperscript{4}}
\DeclareUnicodeCharacter{2075}{\textsuperscript{5}}
\DeclareUnicodeCharacter{2076}{\textsuperscript{6}}
\DeclareUnicodeCharacter{2077}{\textsuperscript{7}}
\DeclareUnicodeCharacter{2078}{\textsuperscript{8}}
\DeclareUnicodeCharacter{2079}{\textsuperscript{9}}
\DeclareUnicodeCharacter{02C2}{<}
\DeclareUnicodeCharacter{2033}{\relax\ifmmode '' \else $''$\fi}
\DeclareUnicodeCharacter{2034}{\relax\ifmmode ''' \else $'''$\fi}
\DeclareUnicodeCharacter{2026}{\relax\ifmmode … \else $\ldots$\fi}
\DeclareUnicodeCharacter{0229}{\c{e}}
\DeclareUnicodeCharacter{016F}{\r{u}}
\DeclareUnicodeCharacter{127}{\relax\ifmmode\rm\hbar\else $\rm\hbar$\fi}
\DeclareUnicodeCharacter{3AC}{\relax\ifmmode\acute{\alpha}\else $\acute{\alpha}$\fi}
\DeclareUnicodeCharacter{3AD}{\relax\ifmmode\acute{\varepsilon}\else $\acute{\varepsilon}$\fi}
\DeclareUnicodeCharacter{3AE}{\relax\ifmmode\acute{\eta}\else $\acute{\eta}$\fi}
\DeclareUnicodeCharacter{3AF}{\relax\ifmmode\acute{\iota}\else $\acute{\iota}$\fi}
\DeclareUnicodeCharacter{3CC}{\relax\ifmmode\acute{o}\else $\acute{o}$\fi}
\DeclareUnicodeCharacter{3CD}{\relax\ifmmode\acute{\upsilon}\else $\acute{\upsilon}$\fi}
\DeclareUnicodeCharacter{3CE}{\relax\ifmmode\acute{\omega}\else $\acute{\omega}$\fi}
\DeclareUnicodeCharacter{391}{A}
\DeclareUnicodeCharacter{392}{B}
\DeclareUnicodeCharacter{395}{E}
\DeclareUnicodeCharacter{396}{Z}
\DeclareUnicodeCharacter{397}{H}
\DeclareUnicodeCharacter{399}{I}
\DeclareUnicodeCharacter{39A}{K}
\DeclareUnicodeCharacter{39C}{M}
\DeclareUnicodeCharacter{39D}{N}
\DeclareUnicodeCharacter{39F}{O}
\DeclareUnicodeCharacter{3A1}{P}
\DeclareUnicodeCharacter{3A4}{T}
\DeclareUnicodeCharacter{3A7}{X}

\DeclareUnicodeCharacter{27E6}{\relax\ifmmode \llbracket \else $\llbracket$\fi}
\DeclareUnicodeCharacter{27E7}{\relax\ifmmode \rrbracket \else $\rrbracket$\fi}

\DeclareUnicodeCharacter{1D434}{\relax\ifmmode A \else $A$\fi}
\DeclareUnicodeCharacter{1D435}{\relax\ifmmode B \else $B$\fi}
\DeclareUnicodeCharacter{1D436}{\relax\ifmmode C \else $C$\fi}
\DeclareUnicodeCharacter{1D437}{\relax\ifmmode D \else $D$\fi}
\DeclareUnicodeCharacter{1D438}{\relax\ifmmode E \else $E$\fi}
\DeclareUnicodeCharacter{1D439}{\relax\ifmmode F \else $F$\fi}
\DeclareUnicodeCharacter{1D43A}{\relax\ifmmode G \else $G$\fi}
\DeclareUnicodeCharacter{1D43B}{\relax\ifmmode H \else $H$\fi}
\DeclareUnicodeCharacter{1D43C}{\relax\ifmmode I \else $I$\fi}
\DeclareUnicodeCharacter{1D43D}{\relax\ifmmode J \else $J$\fi}
\DeclareUnicodeCharacter{1D43E}{\relax\ifmmode K \else $K$\fi}
\DeclareUnicodeCharacter{1D43F}{\relax\ifmmode L \else $L$\fi}
\DeclareUnicodeCharacter{1D440}{\relax\ifmmode M \else $M$\fi}
\DeclareUnicodeCharacter{1D441}{\relax\ifmmode N \else $N$\fi}
\DeclareUnicodeCharacter{1D442}{\relax\ifmmode O \else $O$\fi}
\DeclareUnicodeCharacter{1D443}{\relax\ifmmode P \else $P$\fi}
\DeclareUnicodeCharacter{1D444}{\relax\ifmmode Q \else $Q$\fi}
\DeclareUnicodeCharacter{1D445}{\relax\ifmmode R \else $R$\fi}
\DeclareUnicodeCharacter{1D446}{\relax\ifmmode S \else $S$\fi}
\DeclareUnicodeCharacter{1D447}{\relax\ifmmode T \else $T$\fi}
\DeclareUnicodeCharacter{1D448}{\relax\ifmmode U \else $U$\fi}
\DeclareUnicodeCharacter{1D449}{\relax\ifmmode V \else $V$\fi}
\DeclareUnicodeCharacter{1D44A}{\relax\ifmmode W \else $W$\fi}
\DeclareUnicodeCharacter{1D44B}{\relax\ifmmode X \else $X$\fi}
\DeclareUnicodeCharacter{1D44C}{\relax\ifmmode Y \else $Y$\fi}
\DeclareUnicodeCharacter{1D44D}{\relax\ifmmode Z \else $Z$\fi}
\DeclareUnicodeCharacter{1D44E}{\relax\ifmmode a \else $a$\fi}
\DeclareUnicodeCharacter{1D44F}{\relax\ifmmode b \else $b$\fi}
\DeclareUnicodeCharacter{1D450}{\relax\ifmmode c \else $c$\fi}
\DeclareUnicodeCharacter{1D451}{\relax\ifmmode d \else $d$\fi}
\DeclareUnicodeCharacter{1D452}{\relax\ifmmode e \else $e$\fi}
\DeclareUnicodeCharacter{1D453}{\relax\ifmmode f \else $f$\fi}
\DeclareUnicodeCharacter{1D454}{\relax\ifmmode g \else $g$\fi}
\DeclareUnicodeCharacter{1D456}{\relax\ifmmode i \else $i$\fi}
\DeclareUnicodeCharacter{1D457}{\relax\ifmmode j \else $j$\fi}
\DeclareUnicodeCharacter{1D458}{\relax\ifmmode k \else $k$\fi}
\DeclareUnicodeCharacter{1D459}{\relax\ifmmode l \else $l$\fi}
\DeclareUnicodeCharacter{1D45A}{\relax\ifmmode m \else $m$\fi}
\DeclareUnicodeCharacter{1D45B}{\relax\ifmmode n \else $n$\fi}
\DeclareUnicodeCharacter{1D45C}{\relax\ifmmode o \else $o$\fi}
\DeclareUnicodeCharacter{1D45D}{\relax\ifmmode p \else $p$\fi}
\DeclareUnicodeCharacter{1D45E}{\relax\ifmmode q \else $q$\fi}
\DeclareUnicodeCharacter{1D45F}{\relax\ifmmode r \else $r$\fi}
\DeclareUnicodeCharacter{1D460}{\relax\ifmmode s \else $s$\fi}
\DeclareUnicodeCharacter{1D461}{\relax\ifmmode t \else $t$\fi}
\DeclareUnicodeCharacter{1D462}{\relax\ifmmode u \else $u$\fi}
\DeclareUnicodeCharacter{1D463}{\relax\ifmmode v \else $v$\fi}
\DeclareUnicodeCharacter{1D464}{\relax\ifmmode w \else $w$\fi}
\DeclareUnicodeCharacter{1D465}{\relax\ifmmode x \else $x$\fi}
\DeclareUnicodeCharacter{1D466}{\relax\ifmmode y \else $y$\fi}
\DeclareUnicodeCharacter{1D467}{\relax\ifmmode z \else $z$\fi}

\DeclareUnicodeCharacter{1E67}{\.{\v s}}
\DeclareUnicodeCharacter{1E11}{\relax\ifmmode \c{d} \else $\c{d}$\fi}
\DeclareUnicodeCharacter{1ECB}{\relax\ifmmode \d{i} \else $\d{i}$\fi}
\DeclareUnicodeCharacter{1D8D}{\relax\ifmmode \textlhookx \else $\textlhookx$\fi}
\DeclareUnicodeCharacter{104}{\relax\ifmmode \k{A} \else $\k{A}$\fi}
\DeclareUnicodeCharacter{211E}{\relax\ifmmode \textrecipe \else $\textrecipe$\fi}
\DeclareUnicodeCharacter{29D}{\relax\ifmmode \textctj \else $\textctj$\fi}

\DeclareUnicodeCharacter{1E2E}{\'{\"I}}
\DeclareUnicodeCharacter{23F}{\textrts}

\DeclareUnicodeCharacter{2C73}{\varw}

\DeclareUnicodeCharacter{2127}{\mho}

\DeclareUnicodeCharacter{28C}{\textturnv}
\DeclareUnicodeCharacter{252}{\textturnscripta}
\DeclareUnicodeCharacter{259}{\schwa}
\DeclareUnicodeCharacter{25B}{\m{e}}
\DeclareUnicodeCharacter{266}{\m{h}}
\DeclareUnicodeCharacter{127}{\B{h}}
\DeclareUnicodeCharacter{27E}{\textfishhookr}
\DeclareUnicodeCharacter{281}{\textinvscr}

\newcolumntype{Y}{>{\centering\arraybackslash}X}

\continuouspages{yes}
\firstpage{1} 
\pubvolume{1}
\issuenum{1}
\articlenumber{12345}
\pubyear{2025}
\copyrightyear{2025}
\datereceived{23 May 2025}
\dateaccepted{11 July 2025}
\dateonlinefirst{}
\datepublished{}

\Title{Proactive Disentangled Modeling of Trigger–Object Pairings for Backdoor Defense}

\Author{Kyle Stein\textsuperscript{1}, Andrew A. Mahyari\textsuperscript{1,2}, Guillermo Francia, III\textsuperscript{3} and Eman El-Sheikh\textsuperscript{3}}

\AuthorNames{Kyle Stein, Andrew A. Mahyari, Guillermo Francia, III and Eman El-Sheikh}

\address{%
\textsuperscript{1}Department of Intelligent Systems and Robotics, University of West Florida, Pensacola, FL, 32514, USA

\textsuperscript{2}Florida Institute For Human and Machine Cognition (IHMC), Pensacola, FL, 32502, USA

\textsuperscript{3}Center for Cybersecurity, University of West Florida, Pensacola, FL, 32502, USA
}

\corres{Corresponding Author: Kyle Stein. Email: ks209@students.uwf.edu}

\abstract{Deep neural networks (DNNs) and generative AI (GenAI) are increasingly vulnerable to backdoor attacks, where adversaries embed triggers into inputs to cause models to misclassify or misinterpret target labels. Beyond traditional single-trigger scenarios, attackers may inject multiple triggers across various object classes, forming unseen backdoor-object configurations that evade standard detection pipelines. In this paper, we introduce \textbf{DBOM} (\textbf{D}isentangled \textbf{B}ackdoor-\textbf{O}bject \textbf{M}odeling), a proactive framework that leverages structured disentanglement to identify and neutralize both seen and unseen backdoor threats at the dataset level. \textcolor{black}{Specifically, DBOM factorizes input image representations by modeling triggers and objects as independent primitives in the embedding space through the use of Vision-Language Models (VLMs). By leveraging the frozen, pre-trained encoders of VLMs, our approach decomposes the latent representations into distinct components through a learnable visual prompt repository and prompt prefix tuning, ensuring that the relationships between triggers and objects are explicitly captured. To separate trigger and object representations in the visual prompt repository, we introduce the trigger--object separation and diversity losses that aids in disentangling trigger and object visual features. Next, by aligning image features with feature decomposition and fusion, as well as learned contextual prompt tokens in a shared multimodal space, DBOM enables zero-shot generalization to novel trigger-object pairings that were unseen during training, thereby offering deeper insights into adversarial attack patterns.} Experimental results on CIFAR-10 and GTSRB demonstrate that DBOM robustly detects poisoned images prior to downstream training, significantly enhancing the security of DNN training pipelines. }


\begin{document}
\hypersetup{hidelinks}

\section{Introduction} \label{sect:s1}
As deep neural networks (DNNs) become more prevalent in applications such as natural language processing \cite{devlin2018bert, chowdhary2020natural, galassi2020attention} and object classification \cite{li2022mvitv2, gu2018recent, snell2017prototypical}, they are increasingly being targeted by sophisticated security threats \cite{akhtar2018threat, schwinn2023adversarial}. The rise of generative AI \cite{radford2021learning, goodfellow2020generative, rombach2022high} has enabled the large-scale creation of datasets sourced from online repositories. Although these datasets improve model robustness, they often bypass rigorous vetting, making them vulnerable to backdoor attacks \cite{gu2019badnets, liu2018trojaning, li2020symmetry, nguyen2021wanet}. Such attacks embed hidden triggers in training samples, causing models to misclassify inputs containing the trigger, for example, altering a stop sign’s label to a speed limit sign.

\begin{figure}[t!]
    \centering
    \includegraphics[width=0.6\columnwidth]{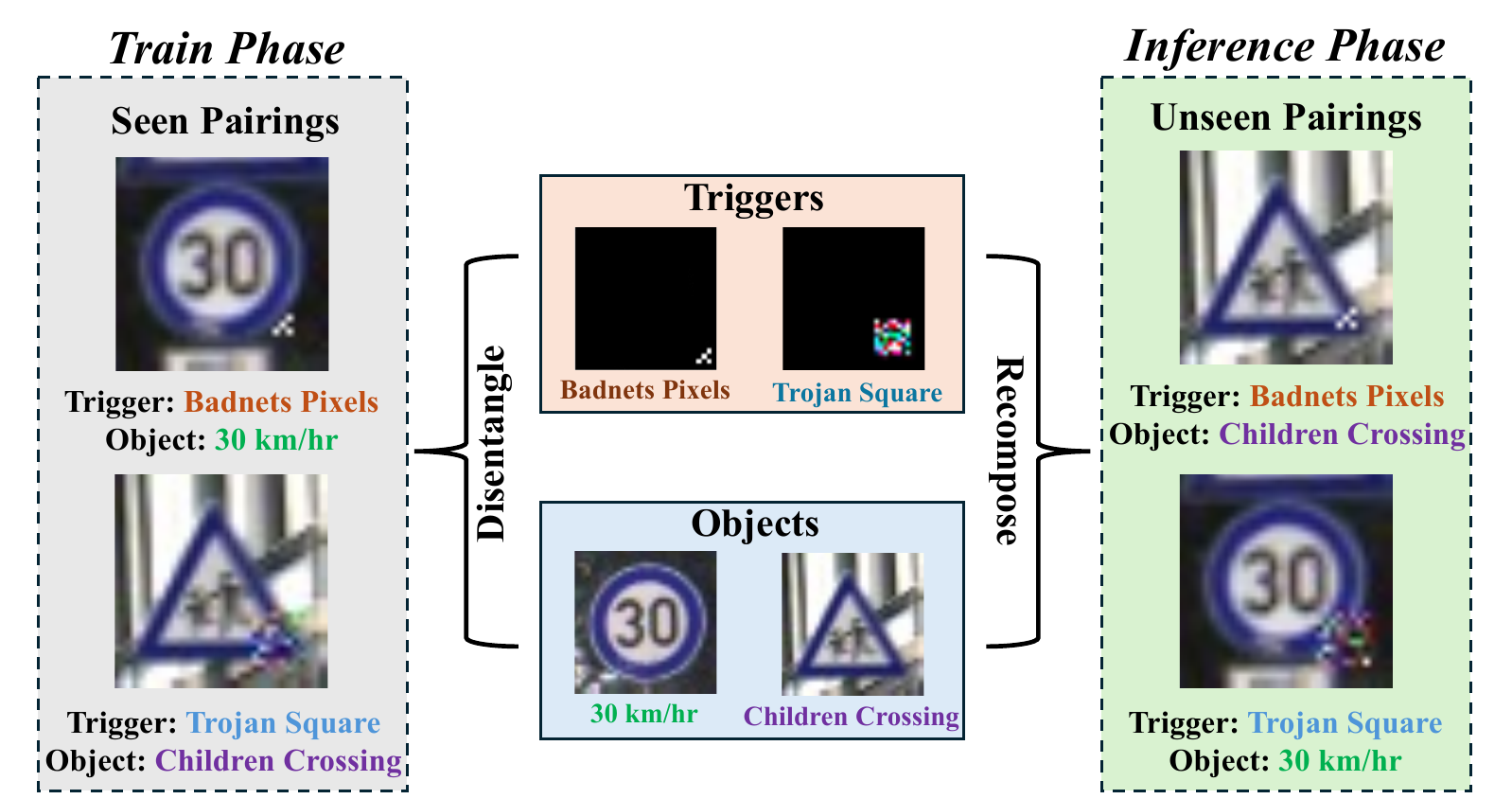} 
    \caption{{Overview of our disentangling process for trigger--object pairings. During training, the system learns separate representations of triggers and objects. By factorizing these components, the model can generalize to unseen trigger--object configurations, although they were never observed together during training.}}
    \label{fig:Disentangle}
\end{figure}

Recent work has focused on identifying backdoored samples in pre-trained infected models \cite{niu2024bdetclip,guo2020meets,yue2022effective,zhu2023enhancing}, but less attention has been given to proactively scanning training data for suspicious triggers before the final model is trained. This lack of focus on the dataset creation phase represents a significant gap in input-level backdoor defense strategies \cite{peri2020deep, kantaros2021real, wen2024holmes, stein2024proactive}. Malicious triggers can be embedded in training samples well before the model is exposed to them, undermining the integrity of the entire training process. Addressing this stage early in the pipeline not only prevents contaminated data from infiltrating the training process, but also reduces the computational costs associated with post-training purification efforts \cite{karim2024augmented, xu2024pad}. Lastly, proactively analyzing the dataset offers deeper insights into the adversarial logic behind these backdoors, specifically how triggers interact with objects and how attackers strategically embed them to exploit vulnerabilities. 

Although existing defenses can detect single or multiple backdoor triggers in a compromised data set \cite{chen2018detecting, tran2018spectral, hayase2021spectre, tang2021demon, ma2022beatrix}, they remain strictly trigger-centric, where flagged samples are discarded, and images of objects classes bearing those triggers are ignored. This removes valuable co-occurrence information into how specific triggers map onto particular objects, which could expose systematic attacker strategies. In realistic many-to-many attack scenarios \cite{li2024shortcuts}, where adversaries plant various triggers across a wide range of object categories, a trigger-only approach would fail to recognize novel trigger-object combinations outside of its training set of known trigger--object pairings. For instance, assume a square-patch trigger is only ever seen on stop signs and a pixel-noise trigger only on speed-limit signs. If an attacker then applies that same square patch to yield signs or the pixel noise to pedestrian-crossing signs (pairings never observed before) those trigger-centric detectors may sharply degrade in performance, since they do not explicitly model which object the trigger appears on. By contrast, a co-occurrence-aware model that simultaneously identifies both triggers and object classes preserves the relational context between adversarial triggers and their targets. Rather than excluding compromised samples, this approach leverages modular relationships to learn comprehensive backdoor patterns and infer previously unseen trigger–object combinations. As a result, the model can accurately recognize the underlying object despite the presence of a trigger, integrate attacked examples into both training and inference workflows, and reduce false positives by distinguishing benign from malicious features. Moreover, modeling trigger–object relationships provides deeper forensic insights into attacker tactics, enabling dynamic update strategies that proactively defends models against evolving many-to-many backdoor attacks. Overall, we can summarize that existing input-level defenses in current state-of-the-art (SOA) attack scenarios remain strictly trigger-centric, where: (1) they identify and discard adversarial samples, losing the underlying object semantics and missing the opportunity to reveal adversarial strategies, (2) do not focus on concurrently identifying triggers and the associated object class, and (3) fail to generalize to novel trigger-object pairings.

To address these gaps, we present Disentangled Backdoor-Object Modeling (\textbf{DBOM}), a proactive framework based on VLMs and prompt tuning \cite{radford2021learning}, designed to identify and isolate unseen backdoor-object configurations. Instead of inspecting a potentially compromised model, this approach focuses on learning trigger-object configurations within web-scraped training images before they are ever fed into a downstream model. Our method surpasses current SOA pre-training defense algorithms by detecting not only the types of backdoor triggers in compromised datasets, but also the underlying objects they target, thereby capturing the adversarial logic behind these malicious trigger–object pairings. Here, we define a trigger as the backdoor attack pattern embedded into an image and an object as the benign semantic class being manipulated. DBOM then factorizes these two primitives into independent embeddings (Figure~\ref{fig:Disentangle}), enabling modular representations of trigger–object configurations \cite{lake2014towards}. Furthermore, by capturing the relationship among triggers and objects during training, \textbf{previously unseen trigger-object pairings can be detected during inference}, a problem traditional single-trigger detection pipelines overlook. The contributions of our approach are as follows:

\begin{itemize}
\item  We introduce DBOM, a novel end-to-end disentangled representation learning framework that separates triggers and objects into independent latent visual primitives. By leveraging cross-modal attention for structured latent decomposition, DBOM aims to learn each trigger pattern and each object class in isolation. At inference, it recomposes these known trigger and object embeddings to recognize combinations never seen during training, achieving zero-shot generalization over trigger–object pairings and resulting in a robust method against adaptive backdoor strategies.

\item Our approach incorporates a dual-branch module that features a learnable visual prompt repository along with a dynamic soft prompt prefix adapter for prompt tuning. The use of a learnable visual prompt repository allows us to capture primitive-specific features for both triggers and objects, aiding in feature disentanglement. Furthermore, dynamically tuning text prompt representations based on image content, our module enhances the semantic context of each sample and improves the separation between trigger and object features. This design allows the framework to capture diverse trigger patterns across multiple object classes, overcoming the limitations of conventional defenses that assume a single, static trigger per dataset.

\item By integrating a proactive backdoor detection mechanism into the data curation process, DBOM identifies unseen backdoor-object attacks before downstream model training begins. A composite loss function that minimizes cross-entropy, disentanglement, and prompt alignment losses together ensures that poisoned samples are identified and isolated for removal from the dataset. 
\end{itemize}

\section{Related Work}
\label{sec:related_work}
\noindent\textbf{Disentanglement} involves separating visual primitives of images into independent components \cite{Tong_2019_CVPR, chen2021semantics, li2021generalized, hao2023learning, stein2025visual}. A central strategy for addressing this task is to train models that learn these independent components and recombine them in novel ways, thereby enabling the flexible recognition of previously unseen trigger--object pairings. Li et al.~\cite{li2020symmetry} apply symmetry and group theory to model primitive relationships, introducing a novel distance function. A Siamese Contrastive Embedding Network (SCEN)~\cite{li2022siamese} embeds visual features into a contrastive space to separately model primitive diversity. A retrieval-augmented approach improves recognition of unseen primitive component pairings by retrieving and refining representations~\cite{jing2024retrieval}. Recent methods integrate vision-language models (VLMs) such as CLIP~\cite{radford2021learning} to enhance the recognition of structured relationships between the underlying nature of images and text prompts. Compositional Soft Prompting (CSP)~\cite{nayak2022learning} utilizes a static prompt prefix alongside learned primitive embeddings, with predictions based on cosine similarity between text and image features. Later works remove the static prefix, making the entire prompt learnable~\cite{xu2024gipcol, lu2023decomposed}. In the context of DBOM, disentangling triggers and objects allows our model to factor visual embeddings into two primitive subspaces: one that captures adversarial trigger patterns and one that encodes the class object semantics. Once these primitives are learned, unseen trigger-object pairings can be inferred upon during testing.

\vspace{1mm}
\noindent\textbf{Backdoor Attacks} became prominent with the introduction of Badnets \cite{gu2019badnets}. Badnets demonstrated how adversaries can embed backdoors into DNNs by poisoning the training data with trigger-patterned images, such as a single white square or pixelated patterns, to misclassify inputs. Liu et.al. \cite{liu2018trojaning} introduced trojaning attacks, which differ from Badnets, by reverse-engineering neuron activations to generate adversarial triggers that maximize activations in specific neurons. Li et.al. \cite{li2020invisible} explored techniques to make triggers more covert to detection by implementing steganographic embedding, where backdoor triggers are hidden within images at a pixel level. Recent backdoor attacks include Wanet \cite{nguyen2021wanet}, a warping-based trigger, which introduces imperceptible image distortions as triggers instead of traditional noise perturbations.

\begin{figure*}[t!]
    \centering
    \includegraphics[width=\textwidth, height=0.25\textheight, keepaspectratio]{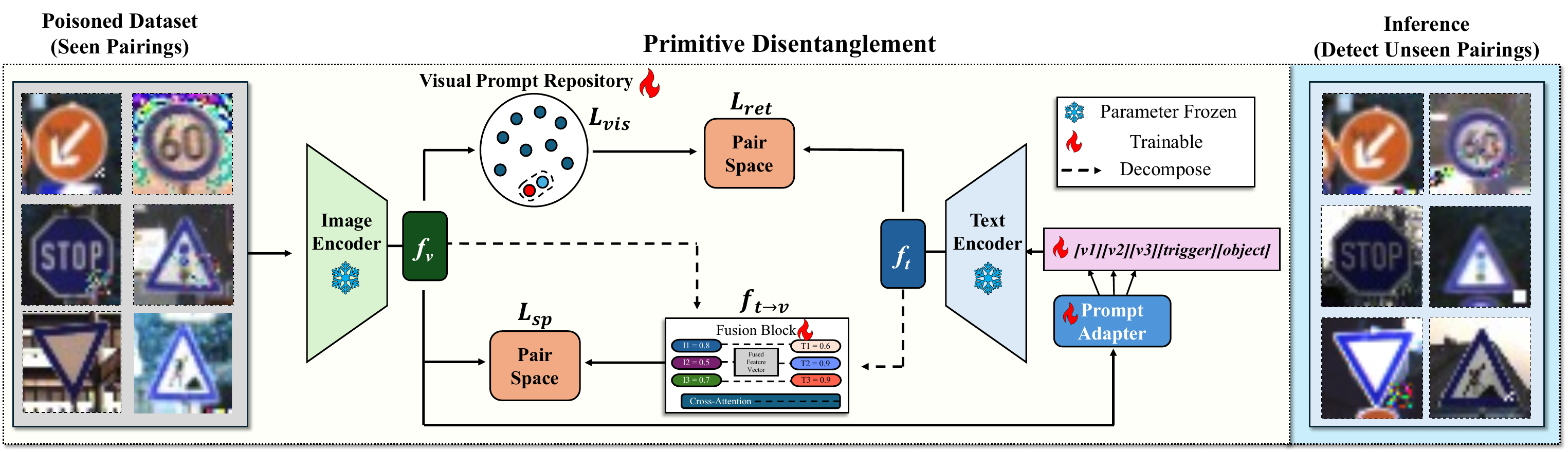} 
    \caption{DBOM utilizes a visual prompt repository and a similarity-based retrieval mechanism to detect unseen backdoor trigger-object representations through the use of CLIP's pre-trained visual and textual encoders. During training, each image retrieves visual prompts from the repository, shifts a learnable text prefix with a prompt adapter, and fuses decomposed image–text features via cross-attention. During inference, the framework again retrieves the top visual prompts, shifts the text prompt for each new image, and computes similarity scores to pinpoint unseen trigger-object pairings. Lastly, in separate pair spaces, the logits are computed by comparing the fused image–text features with the visual features from the frozen visual encoder, as well as the selected visual prompts and the text features from the frozen text encoder. The highest-scoring trigger–object pair is then selected as the predicted configuration. By detecting malicious seen and unseen configurations in this way, DBOM identifies backdoored configurations and isolates them for removal prior to downstream model training.}
    \label{fig:MainFig}
\end{figure*}

\vspace{1mm}
\noindent\textbf{Backdoor Defenses} mostly operate in the adversarial machine learning life-cycle at the model level, leaving the dataset vetting process largely unexplored \cite{wu2023defenses}. Several works attempt to filter adversarial images before training \cite{tang2021demon, kantaros2021real, peri2020deep, wen2024holmes}, but these rely on detecting known trigger-object configurations and fail to generalize to unseen pairings. VisionGuard \cite{kantaros2021real} compares the softmax outputs of original and transformed images using metrics like the Kullback–Leibler divergence to detect attacks without altering the target network. Deep k-NN \cite{peri2020deep} leverages deep feature space clustering and k-nearest neighbor voting to detect and remove poisoned images from the training set prior to downstream model training. HOLMES \cite{wen2024holmes} employs multiple external detectors trained on both dedicated labels and top-k logits to capture subtle differences between benign and adversarial inputs. Traditional backdoor defenses assume a compromised model and attempt to mitigate attacks post-training \cite{zhu2023enhancing, guo2020meets, yue2022effective}. However, \uline{these techniques reactively address attacks after deployment by cleaning the model, whereas our approach proactively filters poisoned images before they enter the downstream training pipeline, preventing backdoor contamination at its source.} Furthermore, these methods overlook the opportunity to identify unseen trigger--object configurations that were not seen in their model training, which is addressed in this paper. 


\vspace{-3mm}
\section{Preliminaries and Insights}
\subsection{Trigger-Object Representation}
\vspace{-1mm}
We define a backdoor configuration as a pairing of a \emph{trigger} and an \emph{object}, where the trigger serves as the adversarial modification and the object represents the underlying semantic class being targeted (e.g., “stop sign,” “yield sign,” “airplane”). Let $T$ be the set of all possible triggers, and $O$ be the set of object categories, where $T = \{t_0, t_1, \dots, t_n\}$ and $O = \{o_0, o_1, \dots, o_m\}$. The complete set of potential trigger–object pairings is given by $P = T \times O$, where each pair $(t, o) \in P$ corresponds to a unique backdoor attack configuration. These pairings can be categorized into two groups: (1) \textit{seen pairings} ($P_s$), which are explicitly observed during training, and (2) \textit{unseen pairings} ($P_u$), which do not appear in the training set but may still be encountered during deployment. These subsets are disjoint ($P_s \cap P_u = \emptyset$) and together form the complete space of possible attack configurations ($P_s \cup P_u = P$). During evaluation, test samples are drawn from a predefined set $P_{\text{test}} \subseteq P$, which contains both seen and unseen pairings. The objective of our approach is to learn a function $f: X \to P_{\text{test}}$, where $X$ represents the input space of images containing these trigger–object configurations. The function $f$ is designed to map an image to its corresponding attack configuration, enabling generalization to \textbf{unseen trigger–object pairs} that were not part of the training distribution. Furthermore, we note that the goal of this paper is not to train an infected model or defend against attacked models, but to detect backdoored images before downstream model training begins.

\vspace{-2mm}
\subsection{Threat Model and Defender Goals}
\vspace{-2mm}
\noindent\textbf{Threat Model.} We assume an adversary injects backdoor attacks based on trigger--object pairings into a web-scraped or publicly available dataset used for training a downstream DNN. The goal is to cause the model to misclassify inputs containing triggers into a target label while maintaining normal classification on clean images. Since large datasets are rarely vetted on a per-sample basis, malicious samples blend easily with clean data. Furthermore, attackers can escalate this threat by injecting multiple triggers across different classes, including novel, unseen trigger–object pairings, so that conventional defenses which expect a single static trigger fail to detect them. Consequently, the compromised data is used in downstream training, embedding hidden adversarial behaviors into the final model. 

\vspace{1mm}
\noindent\textbf{Defender’s Goal.} The defender's goal is to \textit{identify backdoored images prior to downstream model training}, ensuring they are isolated while minimizing the misclassification of clean images. Given a potentially poisoned dataset that contains several triggers–object configurations, the defender must distinguish legitimate images from those carrying triggers. Furthermore, by concurrently identifying both the trigger and the underlying object, the defender learns vital information into the adversary's strategies. Moreover, separating the adversarial trigger from the underlying object enables the recovery of correct object semantics in backdoored samples, eliminating the need to discard these adversarial samples from training or inference. 


\section{Proposed Framework}
\label{sec:methodology}
DBOM leverages CLIP as its backbone by freezing its pre-trained visual and text encoders. Let $f_\theta(\cdot)$ denote the CLIP image encoder and $g_\phi(\cdot)$ denote the CLIP text encoder. Given an input image $x_i$, the image encoder extracts visual features $f_v = f_\theta(x_i) \in \mathbb{R}^d$, which serve two purposes: (i) they are used to retrieve the most relevant visual prompts from a learnable repository, and (ii) they provide the bias for shifting a set of learnable prefix text tokens \texttt{[v1][v2][v3]} via a prompt adapter network. Unlike fixed prefix templates (i.e., \texttt{a photo of}), our approach employs \textit{prompt tuning}, a technique where these prefix tokens are treated as learnable parameters and optimized end-to-end to capture task-specific context for each image. This allows the text prompt to be tailored to the visual content of each image, promoting the alignment between visual and textual modalities. The shifted prefix is then appended to the trigger and object word embeddings to form the final prompt $t_i$, which is processed by the text encoder to produce text features $f_t = g_\phi(t_i) \in \mathbb{R}^{768}$. Lastly, $f_v$ and $f_t$ are decomposed and fused, and their joint representation is mapped into a separate pair space where the similarity between the image and fused features helps determine the final trigger-object prediction. Figure \ref{fig:MainFig} displays the overall architecture of the proposed approach.

\subsection{Visual Prompt Repository}
The visual prompt repository comprises a collection of $M$ learnable visual prompts $\{\mathbf{P}_i\}_{i=1}^{M}$, with each prompt $\mathbf{P}_i \in \mathbb{R}^{l \times d}$ paired with a learnable key $\mathbf{a}_i \in \mathbb{R}^{d}$. These prompts capture high-level visual semantics and are refined during training. For a given image, cosine similarity is computed between the normalized image features $f_v$ and each normalized key. Based on the similarity scores, the two most similar prompts are selected. One is intended to align with the image’s \textbf{trigger} and the other with the \textbf{object}. To enforce this specialization, we introduce two auxiliary losses. The \emph{trigger-object separation loss} is formulated as:

\begin{equation}
\scriptsize
\mathcal{L}_{\text{sep}}
= -\frac{1}{N}\!\sum_{i=1}^N
\log \Biggl(
\frac{\exp\!\bigl(\cos(f_v^{(i)}, \mathbf{a}_{\text{trig}}^{(i)})\bigr)}{
\exp\!\bigl(\cos(f_v^{(i)}, \mathbf{a}_{\text{trig}}^{(i)})\bigr) + 
\exp\!\bigl(\cos(f_v^{(i)}, \mathbf{a}_{\text{obj}}^{(i)})\bigr)
}
\Biggr).
\scriptsize
\end{equation}

\noindent
Because our primary objective is to accurately flag backdoored images, the loss function prioritizes the trigger key by encouraging it to achieve a higher similarity score than the object key, with the object serving as complementary context for the image. The \emph{visual prompt diversity loss} is defined as:

\begin{equation}
\mathcal{L}_{\text{div}} 
= \frac{1}{N} \sum_{i=1}^{N} 
\max\!\Bigl(0,\, m \;-\; \cos\bigl(\mathbf{a}_{\text{trig}}^{(i)}, \mathbf{a}_{\text{obj}}^{(i)}\bigr)\Bigr),
\end{equation}

\noindent where \(m=0.5\) is a fixed margin. This term penalizes any excessive similarity between the retrieved trigger and object visual prompts, thereby promoting disentangled features for more distinct representations \cite{oord2018representation}. Combining these terms yields:

\begin{equation}
\mathcal{L}_{\text{vis}} = \mathcal{L}_{\text{sep}} + \mathcal{L}_{\text{div}},
\end{equation}

\noindent which guides the prompts to distinctly capture trigger and object characteristics. During training, the visual prompt repository is updated end-to-end with \(\mathcal{L}_{\text{vis}}\). This ensures that the repository vectors are not static but are continuously refined to distinguish between trigger and object features. The final representation of the retrieved visual prompts can be denoted by $f_{ret}$.

\subsection{Dynamic Prefix Adapter}
Traditional prompt tuning approaches \cite{zhou2022conditional, nayak2022learning, radford2021learning} use a fixed soft prompt prefix, where a sequence such as \texttt{[trigger][object]} is appended with an initialized phrase \texttt{a photo of}. This means that the same prefix is applied to every sample, regardless of the unique characteristics of the trigger or object in the image. This prefix rigidity can hinder the system’s ability to accurately distinguish between different trigger--object pairs. Motivated by the work in \cite{zhou2022conditional}, we propose an adaptive prompt network module that dynamically adjusts the learnable prefix tokens based on the visual content of the input image. This has been shown to transfer the frozen backbone’s generalization power to entirely new tasks with very few labeled examples \cite{zhou2022learning, zhou2022conditional, khattak2023maple}. 

Specifically, the prompt adapter utilizes the image features \(f_v\) to compute a bias term that is added to the base prompt tokens, thus tailoring the prompt to each individual sample. Besides, by dynamically shifting the soft‐prompt prefix based on each image’s visual features, the prompt prefix adapter aligns the text embeddings more closely with the specific trigger and object primitives, which in turn lets the model accurately recombine those known primitives into novel, unseen pairings at inference, improving zero-shot pairing performance. The prompt adapter is implemented as a lightweight neural network defined by:

\begin{equation}
\text{APNet}(f_v) = {\bf W}_2 \cdot \sigma({\bf W}_1 \cdot f_v + {\bf b}_1) + {\bf b}_2,
\end{equation}

\noindent where \(\sigma(\cdot)\) denotes the ReLU activation function, and \({\bf W}_1\), \({\bf W}_2\), \({\bf b}_1\), and \({\bf b}_2\) are trainable parameters. The output, $\varphi(f_v)$, represents the bias added element-wise to the original prompt embeddings $\{\theta_0, \theta_1, \dots, \theta_p\}$ via $\theta'_i = \theta_i + \varphi_i(f_v)$ for $i = 0, \dots, p$. The final text prompt $t_i$ is constructed by appending $\{\theta'_0, \theta'_1, \dots, \theta'_p\}$ with the trigger and object word embeddings, $\theta_t$ and $\theta_o$, respectively. Lastly, $t_i$ is fed into the text encoder to generate the text features $f_t$. 


\subsection{Feature Decomposition and Fusion}
To disentangle and jointly embed the representations of \textbf{triggers} and \textbf{objects} for backdoor detection, we decompose and then fuse the visual features, \( f_v \), and the text features, \( f_t \) \cite{lu2023decomposed}. We first isolate how each trigger and object contributes to the text representation by averaging their respective logits. This decomposition helps the model treat triggers and objects as independent primitives, ensuring that potential backdoor triggers are not blended with the underlying objects during subsequent fusion. During training, we explicitly supervise these decomposed features to capture the semantics of each trigger and object class.

Formally, we compute the trigger and object probabilities as follows:

\begin{equation}
p(y = t \mid x; \theta) = \frac{\exp(f_v \cdot f_t)}{\sum\limits_{\bar{t} \in \mathcal{T}} \exp(f_v \cdot f_t)},
\end{equation}

\begin{equation}
p(y = o \mid x; \theta) = \frac{\exp(f_v \cdot f_t)}{\sum\limits_{\bar{o} \in \mathcal{O}} \exp(f_v \cdot f_t)},
\end{equation}

\noindent where \(\mathcal{T}\) is the set of possible triggers, \(\mathcal{O}\) is the set of possible objects, and \(\theta\) denotes the learnable parameters. We then optimize cross-entropy losses for the trigger (\(\mathcal{L}_{\text{tri}}\)) and object (\(\mathcal{L}_{\text{obj}}\)) predictions:

\begin{equation}
\mathcal{L}_{\text{tri}} 
= - \frac{1}{| \mathcal{T} |} 
\sum_{(x, y) \in \mathcal{P}^s} 
\log \bigl( p\bigl( y=(t) \mid x; \theta \bigr) \bigr),
\end{equation}

\begin{equation}
\mathcal{L}_{\text{obj}} 
= - \frac{1}{| \mathcal{O} |} 
\sum_{(x, y) \in \mathcal{P}^s} 
\log \bigl( p\bigl( y=(o) \mid x; \theta \bigr) \bigr),
\end{equation}
where \(\mathcal{P}^s\) denotes the set of \textbf{seen} triggers--object pairings.

Next, \( f_v \) and \( f_t \) are fused with a cross-attention mechanism that aligns the image and text features within a joint embedding space. Specifically, we define the query \( Q \) from \( f_t \), and the key \( K \) and value \( V \) from \( f_v \). The query identifies the textual aspects that need to be emphasized in the visual representation; the key--value pairs in the visual space highlight regions or features corresponding to each textual element:

\begin{equation}
\text{Attention}(Q, K, V) = 
\text{softmax} \Bigl( \frac{Q K^{T}}{\sqrt{d}} \Bigr)\, V,
\end{equation}

\noindent where \( d \) is the feature dimensionality. The result of this cross-attention is \( f_{t \rightarrow v} \), a fused representation that integrates the textual context of the triggers and objects with the corresponding visual features.

\subsection{Training and Inference}
Our framework trains in two main stages: we first adapt the soft prompt so that the fused features \(f_{t \rightarrow v}\) correctly capture the target trigger--object pairings, and then we ensure the textual representation \(f_t\) is consistent with the retrieved visual prompt. We compute the probability of a trigger--object pair \(\bigl(t, o\bigr)\) by comparing the image feature \(f_v\) to the fused representation \(f_{t \rightarrow v}\):

\begin{equation}
p_{\text{sp}}\bigl( y=(t, o) \mid x; \theta \bigr)
= \frac{\exp\bigl( f_v \cdot f_{t \rightarrow v} \bigr)}%
       {\sum_{(t',o') \in \mathcal{P}^s} \exp\bigl( f_v \cdot f_{t \rightarrow v} \bigr)}.
\end{equation}

\noindent Minimizing the cross-entropy over these probabilities yields  the soft prompt alignment loss \(\mathcal{L}_{\text{sp}}\). This encourages the shifted soft prompt to correctly identify  the trigger--object pairs for samples in \(\mathcal{P}^s\). Next, we require that the textual representation \(f_t\) matches the retrieved pairing from the prompt repository. We define:

\begin{equation}
p_{\text{ret}}\bigl( y=(t, o) \mid x; \theta \bigr) 
= \frac{\exp\bigl( f_{ret} \cdot f_t \bigr)}%
       {\sum_{(t',o') \in \mathcal{P}^s} \exp\bigl( f_{ret} \cdot f_t \bigr)}.
\end{equation}

\noindent Minimizing the cross-entropy over these probabilities produces the retrieval alignment loss \(\mathcal{L}_{\text{ret}}\). The total loss is a weighted sum of these components along with the prompt losses:

\begin{equation}
\mathcal{L}_{\text{total}} = \mathcal{L}_{\text{ret}} + \lambda_{\text{tri\_obj}} \Big( \mathcal{L}_{\text{tri}} + \mathcal{L}_{\text{obj}} \Big) + \lambda_{\text{sp}}\, \mathcal{L}_{\text{sp}} + \lambda_{\text{vis}}\mathcal{L}_{\text{vis}}.
\end{equation}

During inference, the learned prompt adapter shifts the prefix tokens, the visual prompts are retrieved and averaged, and the logits are computed based on the similarity between the image and text features in the pair space. The predicted trigger--object text labels are selected by:
\begin{equation}
\hat{y} = \underset{(t, o) \in \mathcal{P}^{test}}{\arg \max} \, p_{\text{sp}}\bigl(y = (t, o) \mid x; \theta\bigr),
\end{equation}
where $\mathcal{P}^{test}$ denotes the set of test trigger--object pairings, which includes seen and unseen configurations, and \(p_{\text{sp}}\) is computed following the same procedure in Eq.\,(10).

\section{Experiments and Results}
\label{sec:experiment}
\subsection{Experimental Setup}
\noindent \textbf{Attacks and Splits.} 
We conduct experiments using two benchmark datasets: CIFAR-10 \cite{krizhevsky2009learning} and GTSRB \cite{stallkamp2012man}. CIFAR-10 contains 50,000 training images and 10,000 test images across 10 object classes, while GTSRB consists of 39,209 training images and 12,630 test images spanning 43 traffic sign classes. Recent studies \cite{kantaros2021real, yao2019latent} have shown that adversaries can place backdoor triggers directly on traffic signs to mislead advanced driver‐assistance and autonomous-driving systems. Therefore, GTSRB provides a practical, safety-critical testbed for evaluating our proposed data-level defense system. To introduce backdoor vulnerabilities, we generate contaminated versions of all clean images using six attack patterns, while retaining the clean images themselves as an individual class. The six widely recognized backdoor attacks which are employed are: Badnets Square (Badnets-SQ) \cite{gu2019badnets}, Badnets Pixels (Badnets-PX) \cite{gu2019badnets}, Trojan Square (Trojan-SQ) \cite{liu2018trojaning}, Trojan Watermark (Trojan-WM) \cite{liu2018trojaning}, $l_2$-inv \cite{li2020invisible}, and $l_0$-inv \cite{li2020invisible}. These attacks encompass a diverse range of backdoor characteristics, including universality, label specificity, and variations in trigger shape, size, and placement. This results in a trigger--object pairing space of 301 unique pairings for GTSRB and 70 pairings for CIFAR-10. 

\noindent\textbf{Implementation Details.} We utilize PyTorch 1.12.1 \cite{paszke2019pytorch} for the implementation of our model. The model is optimized using the Adam optimizer \cite{kingma2014adam} and is trained over 20 epochs on the previously mentioned datasets. Both the image encoder and text encoder are based on the pretrained CLIP ViT-L/14 model, and the entire model is trained and evaluated on a single NVIDIA 2080 Ti GPU. We set $M = 20$ for both GTSRB and CIFAR-10. To assess scalability and accuracy trade-offs, all experiments are implemented with the smaller CLIP variants ViT-B/16 and ViT-B/32, repeating the same training schedule and hyperparameters.

\subsection{Unseen Trigger--Object Evaluation}
This experiment evaluates the performance of DBOM in both the seen (S) and unseen (U) trigger--object pairing scenarios. Specifically, the accuracy for each trigger--object pairing type is measured, assessing both the Attack (trigger) and Object classifications separately. To provide a comprehensive evaluation, we report the Harmonic Mean (HM) of the seen and unseen accuracies, which balances performance across known and novel pairings. In addition, we calculate the area under the curve (AUC), which serves as the primary metric for assessing the overall effectiveness of the model in detecting trigger-object configurations. We compare DBOM's results with CoOP \cite{zhou2022conditional} and CSP \cite{nayak2022learning} since they represent two distinct approaches for leveraging CLIP in modeling triggers and objects as separate primitives in the embedding space. CoOP uses fixed, pre-computed natural language representations for the triggers and objects while learning only a context prompt prefix to condition CLIP. In contrast, CSP learns soft prompts by fine-tuning learnable tokens for triggers and objects, allowing for more adaptive reconfiguration and improved generalization to unseen trigger–object pairings.

\begin{table*}[t!]
  \centering
    \caption{Comparison of backdoor trigger–object identification methods on GTSRB and CIFAR-10. Bold indicates the best results.}
  \small
  \begin{tabularx}{\textwidth}{l|l|*{6}{C}|*{6}{C}}
    \toprule
    \multirow{2}{*}{\textbf{Method}} & \multirow{2}{*}{CLIP Model} & \multicolumn{6}{c|}{\textbf{GTSRB}} & \multicolumn{6}{c}{\textbf{CIFAR-10}} \\
    \cmidrule(lr){3-8} \cmidrule(lr){9-14}
    & & S & U & Att. & Obj. & HM & AUC & S & U & Att. & Obj. & HM & AUC \\
    \midrule
    CoOP \cite{zhou2022conditional}  & ViT-L/14 & 28.26 & 28.95 & 37.26 & 35.59 & 11.59 & 4.95 & 65.64 & 67.81 & 46.31 & 92.69 & 47.47 & 35.67 \\
    CSP \cite{nayak2022learning}     & ViT-L/14 & 57.34 & 77.86 & 65.27 & 76.85 & 51.07 & 38.03 & 70.28 & 77.81 & 63.34 & \textbf{95.28} & 62.23 &  50.42\\
    \midrule
    DBOM (Ours)                           & ViT-B/32 & 92.65 & 93.70 & 98.31 & 87.10 & 88.05 & 85.03 & 92.09 & 93.76 & 98.19 & 87.38 & 86.76 & 84.43 \\
    DBOM (Ours)                          & ViT-B/16 & 93.19 & 95.47 & \textbf{98.63} & 90.32 & 90.21 & 87.86 & 93.40 & 94.90 & 98.31 & 89.51 & 90.22 & 87.37 \\
    DBOM (Ours)                            & ViT-L/14 & \textbf{96.89} & \textbf{96.88} & 98.15 & \textbf{95.00} & \textbf{93.94} & \textbf{92.29} & \textbf{96.90} & \textbf{98.15} & \textbf{98.80} & 95.20 & \textbf{94.19} & \textbf{93.07} \\
    \bottomrule
  \end{tabularx}
  \label{tab:results}
\end{table*}

Table \ref{tab:results} demonstrates that DBOM outperforms the baseline methods across nearly all metrics. DBOM improves AUC over 53\% on GTSRB and nearly 43\% on CIFAR-10. Furthermore, DBOM successfully identifies over 98\% of backdoor triggers on both benchmarks while classifying nearly 95\% of objects in the diverse GTSRB dataset (43 classes) and over 95\% on CIFAR-10 (10 classes). Importantly, the high accuracy observed for unseen trigger-object pairings indicates that our model can detect trigger-object pairings that were \textbf{not encountered} during training. Note that DBOM not only generalizes to unseen trigger–object pairings, it also accurately identifies seen triggers: the “Seen” columns in Table~\ref{tab:results} show over 92 and 96\% accuracy on known trigger patterns. 

Moreover, we report the results of smaller CLIP variants in Table \ref{tab:results} and average run-times across both datasets for each variant in Table \ref{tab:runtime}. We can observe that the ViT-B/32 and ViT-B/16 models run at an average of 2.53 ms and 4.27 ms/image, compared to ViT-L/14's 10.69 ms/image respectively. Importantly, this reduction in compute does not result in a significant drop in accuracy: the ViT-B/32–based DBOM still achieves AUC scores of 85.03\% on GTSRB and 84.43\% on CIFAR-10, while the ViT-B/16 variant increases those figures to 87.86\% and 87.37\%. These findings suggest that our approach can leverage smaller CLIP backbones for real-time deployment without sacrificing the high trigger-object identification performance afforded by the larger variant.

\begin{table}[t]
  \centering
  \small
  \caption{Inference runtime per image on a single NVIDIA 2080 Ti GPU (batch size 64).}
  \label{tab:runtime}
  \begin{tabular}{lc}
    \toprule
    \textbf{CLIP Variant} & \textbf{Inference Time (ms/img)} \\
    \midrule
    ViT-B/32 & 2.53 \\
    ViT-B/16 & 4.27 \\
    ViT-L/14 & 10.69 \\
    \bottomrule
  \end{tabular}
\end{table}

Overall, DBOM's zero-shot generalization capability to novel trigger--object pairings is achieved by leveraging the disentangled representation learning approach, which factors triggers and objects into independent primitives. Although previous methods aim for similar generalization, our visual prompt repository, dynamic prefix adapter, feature decomposition and fusion greatly improve the ability to recombine these learned representations to accurately identify novel trigger-object pairings. Therefore, DBOM offers robust protection against evolving backdoor attack strategies by possessing the ability to identify seen configurations with high accuracy and then leveraging those seen pairings to identify unseen configurations, resulting in an adaptive method that can simultaneously evolve to adversarial strategies.

\subsection{Backdoor Poison Detection Evaluation}
DBOM is compared against conventional pre-training dataset cleaning approaches \cite{peri2020deep, kantaros2021real, wen2024holmes} by simulating a realistic scenario where the poisoning rate is set at 5\%, 10\%, and 15\%, reflecting the poisoning ratios often encountered in web-scraped datasets. Overall accuracy (Acc.) measures the proportion of all images, both clean and poisoned, that are correctly classified. Futhermore, we report the attack recall (Rec.), indicating the percentage of poisoned images that are successfully identified. Additionally, attack precision (Prec.) measures the proportion of images flagged as attacked that are truly poisoned, and the F1 Attack score is the harmonic mean of attack precision and recall. Table~\ref{tab:poison_detection} summarizes the performance of DBOM relative to baseline methods.

\begin{table*}[t!]
  \centering
    \caption{Poison detection evaluation at 5\%, 10\%, and 15\% poisoning levels on CIFAR-10 and GTSRB. Bold indicates the best results for each poisoning rate.}
  \small
  \begin{tabularx}{\textwidth}{l|c|*{4}{C}|*{4}{C}}
    \toprule
    \multirow{2}{*}{\textbf{Method}} & \multirow{2}{*}{\textbf{Poisoning Rate}} 
    & \multicolumn{4}{c|}{\textbf{GTSRB}} & \multicolumn{4}{c}{\textbf{CIFAR-10}} \\
    \cmidrule(lr){3-6} \cmidrule(lr){7-10}
    & & Acc. & Rec. & Prec. & F1 & Acc. & Rec. & Prec. & F1 \\
    \midrule
    \multirow{3}{*}{VisionGuard \cite{kantaros2021real}} 
      & 5\%  & 88.43  & 57.07   & 23.23   & 33.02  & 85.56   & 48.57  & 16.94  & 25.12 \\
      & 10\% & 85.09  & 62.16   & 35.83   & 45.46  & 88.34   & 65.32  & 44.34  & 52.82 \\
      & 15\% & 90.23  & 63.29   & 68.99   & 66.02  & 90.94   & 70.17  & 69.58  & 69.87 \\
    \midrule
    \multirow{3}{*}{Deep \textit{k}-NN \cite{peri2020deep}} 
      & 5\%  & \textbf{99.46} & 89.13   & \textbf{100.0} & 94.25  & 98.81   & 76.19  & \textbf{100.0} & 86.49 \\
      & 10\% & 97.11  & 75.35   & 95.65   & 84.40  & 97.59   & 75.90  & \textbf{100.0} & 86.30 \\
      & 15\% & 94.69  & 64.59   & \textbf{100.0} & 78.48  & 97.45   & 82.95  & \textbf{100.0} & 90.68 \\
    \midrule
    \multirow{3}{*}{HOLMES \cite{wen2024holmes}} 
      & 5\%  & 95.99  & 35.56   & 96.97   & 52.03  & \textbf{99.29} & 80.00  & \textbf{100.0} & 88.89 \\
      & 10\% & 96.91  & 69.81   & \textbf{100.0} & 82.22  & 97.53   & 78.43  & \textbf{100.0} & 87.91 \\
      & 15\% & 93.62  & 57.20   & 99.29   & 72.58  & 97.45   & 83.10  & \textbf{100.0} & 90.77 \\
    \midrule
    \multirow{3}{*}{DBOM (Proposed)}  
      & 5\%  & 98.36  & \textbf{98.49} & 98.83   & \textbf{98.63}  & 97.86   & \textbf{97.23} & 98.86  & \textbf{98.19} \\
      & 10\% & \textbf{98.05} & \textbf{95.52} & 98.21   & \textbf{96.83}  & \textbf{98.80} & \textbf{98.79} & 99.05  & \textbf{98.85} \\
      & 15\% & \textbf{97.86} & \textbf{98.19} & 98.28   & \textbf{98.23}  & \textbf{97.58} & \textbf{97.58} & 98.06  & \textbf{97.71} \\
    \bottomrule
  \end{tabularx}
  \label{tab:poison_detection}
\end{table*}

\begin{figure*}[t!]
    \centering
    \includegraphics[width=\textwidth, height=0.25\textheight, keepaspectratio]{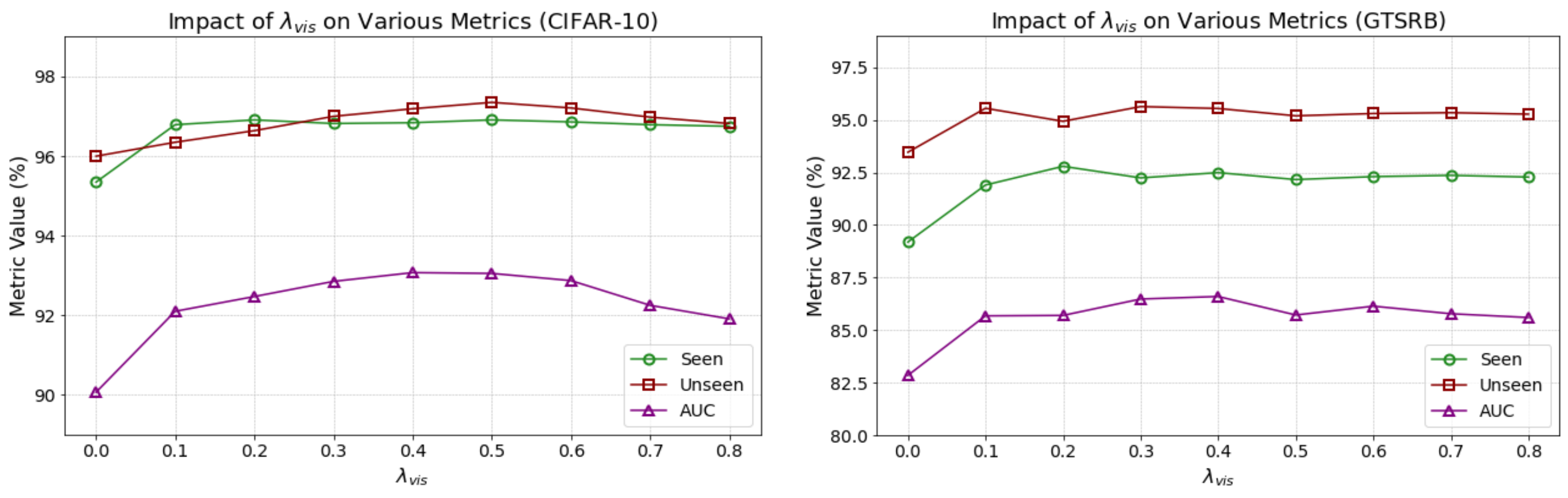} 
    \caption{Impact of $\lambda_{\text{vis}}$ on AUC and seen/unseen accuracy.}
    \label{fig:Metric_Lambda}
\end{figure*}

Evaluation shows that DBOM consistently results in high overall accuracy while keeping the misclassification of clean samples to a minimum. For example, on GTSRB, DBOM achieves overall accuracies of around 98\% with an attack recall consistently exceeding 97\% and F1 scores near 98\% across poisoning rates of 5\%--15\%. Similar trends are observed on CIFAR-10, where overall accuracies are in the range of 97--98\%, and both attack recall and F1 scores remain high. Furthermore, our experimental results reveal an important trade-off between precision and recall. While methods such as Deep \textit{k}-NN and HOLMES achieve near perfect precision, they often suffer from lower attack recall (typically around 75--80\%), leading to significantly lower F1 scores. DBOM's modest decrease in precision is acceptable because missing a poisoned image can be far more harmful than incorrectly flagging a few additional clean images, especially when clean images make up the majority of the dataset. Lastly, unlike existing SOA methods that solely focus on identifying whether an image is backdoored or poisoned, DBOM disentangles each image's representations into primitives to identify both the trigger and the object concurrently, thereby enabling it to detect unseen configurations that were not encountered during training, a crucial improvement over existing SOA methods.

\subsection{Ablation Study}

\noindent \textbf{Impact of \(\lambda_{\text{vis}}\).} We investigate the influence of the visual prompt loss weight, \(\lambda_{\text{vis}}\), on DBOM's ability to disentangle trigger and object features. Recall that the visual prompt loss \(\mathcal{L}_{\text{vis}} = \mathcal{L}_{\text{sep}} + \mathcal{L}_{\text{div}}\) enforces higher similarity for the trigger visual prompt and diversity between the trigger and object visual prompts. Note that when \(\lambda_{\text{vis}}\) = 0.0, the visual prompt loss is removed from the training objective and the model loses supervision to disentangle trigger and object features from the visual prompt repository, although the top two most similar prompts are still selected. 

The results, shown in in Figure \ref{fig:Metric_Lambda}, reveal that at $\lambda_{\text{vis}} = 0.0$, the model achieves the lowest performance across all metrics. As $\lambda_{\text{vis}}$ increases, the supervision provided by the separation and diversity losses leads to improvements in both AUC and unseen accuracy, reaching a peak at $\lambda_{\text{vis}} = 0.5$. This peak indicates that a moderate emphasis on the separation losses most effectively refines the latent representations. Therefore, the model is able to generalize more robustly to unseen backdoor configurations. While selecting the top two prompts from the visual repository yields acceptable performance, incorporating the explicit separation and diversity losses significantly improves overall performance across all metrics. While results on CIFAR-10 show a more stable rise and fall of seen, unseen, and AUC values, the results on GTSRB show more variation over each tested $\lambda_{\text{vis}}$ value. 

\vspace{2mm}
\noindent \textbf{Learnable vs. Static Prefix.} In this experiment, we replace the learnable soft prompt adapter with a static fixed prompt prefix, \texttt{a photo of}, to isolate the influence of a constant prefix context on model performance. Table \ref{tab:static} details the performance improvement across all metrics of the learnable prefix adapter over the fixed prefix. For GTSRB, the learnable prefix leads to a 5.07\% increase in object classification accuracy, AUC 3.31\% and seen accuracy 2.19\%. This improvement is especially significant for object classification, given that GTSRB has a diverse set of 43 classes, making the task more challenging. Similarly, on CIFAR-10, we see a notable 1.59\% increase on unseen pairings, 1.38\% for object classification, and 1.92\% for AUC. The improvements can be attributed to dynamically adjusting the prefix tokens based on each input image's content, leading to better alignment between visual and textual representations and more precise detection. This improves the model's capability to distinguish between triggers and objects, especially when encountering unseen adversarial configurations.  

\begin{table}[t!]
\centering
\caption{Learnable vs. Static Prefix.}
\setlength{\tabcolsep}{4pt} 
\renewcommand{\arraystretch}{1.2} 
\resizebox{0.5\columnwidth}{!}{%
\begin{tabular}{lcccc}
\toprule
\textbf{Method} & \multicolumn{2}{c}{\textbf{GTSRB}} & \multicolumn{2}{c}{\textbf{CIFAR-10}} \\ 
\cmidrule(lr){2-3} \cmidrule(lr){4-5} 
                & ``a photo of" & [v1][v2][v3] & ``a photo of" & [v1][v2][v3] \\ 
\midrule
 Seen & 90.10 &  92.29 (\textcolor{blue}{+2.19}) & 96.75 & 96.90 (\textcolor{blue}{+0.15})\\
 Unseen &  94.92 &  95.54 (\textcolor{blue}{+0.62}) & 95.75 & 97.34 (\textcolor{blue}{+1.59})\\
 Attack & 97.76 &  98.04 (\textcolor{blue}{+0.28})  & 97.74 &  98.80 (\textcolor{blue}{+1.06})\\
 Object & 84.09 &  89.16 (\textcolor{blue}{+5.07})  & 93.82 &  95.20 (\textcolor{blue}{+1.38}) \\
 HM &  86.40 &  87.50 (\textcolor{blue}{+1.10}) & 93.01 & 94.19 (\textcolor{blue}{+1.18}) \\
 AUC  & 83.29 &  86.60 (\textcolor{blue}{+3.31})  & 91.15 & 93.07 (\textcolor{blue}{+1.92})\\
\bottomrule
\end{tabular}%
}
\label{tab:static}
\end{table}

\subsection{Qualitative Analysis}

\begin{figure*}[t!]
    \centering
    \includegraphics[width=\textwidth, height=0.25\textheight, keepaspectratio]{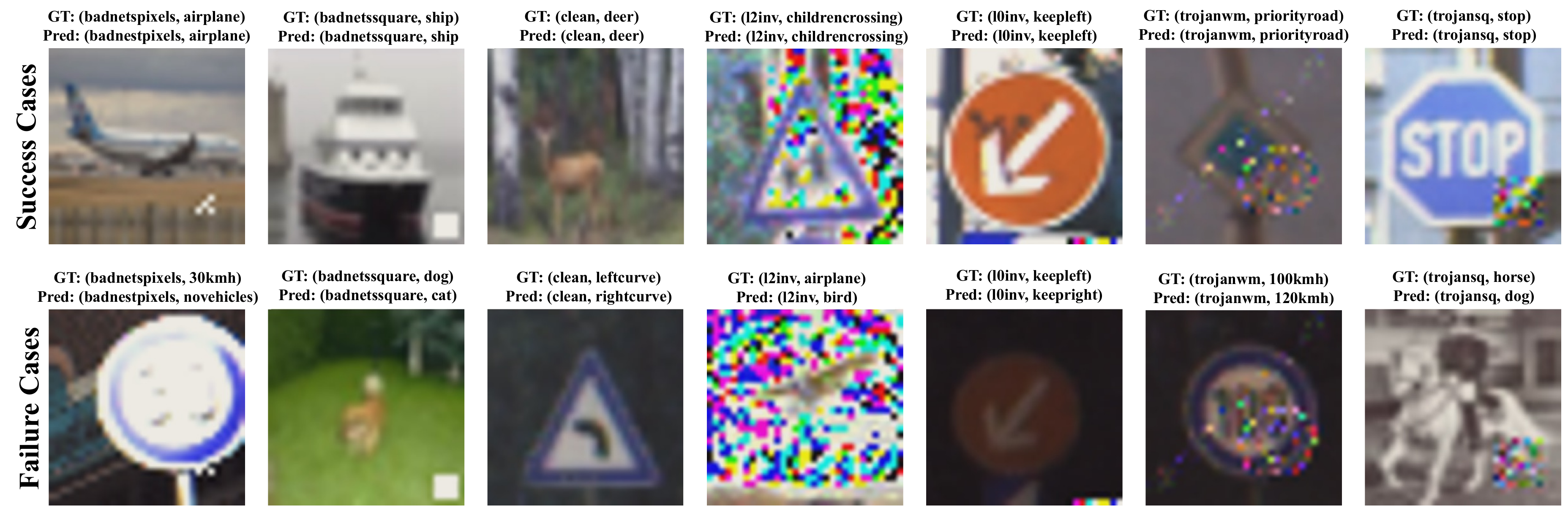} 
    \caption{Ground Truth vs. Prediction of DBOM.}
    \label{fig:Qualitative}
\end{figure*}

Figure \ref{fig:Qualitative} displays randomly selected images from the test set along with the predicted trigger–object pairs and their ground-truth labels. The examples in the top row highlight \textit{successful predictions}, illustrating how our framework can handle diverse triggers, object classes, and varying image quality. Even in blurry or distorted cases, such as the “Priority Road” sign, the model still distinguishes both the trigger and the object accurately.

In contrast, the bottom row depicts \textit{failure cases} where the predicted objects differ from the ground truth (though the triggers are correctly identified). For instance, in the first error image, “30km/h” is misclassified as “No Vehicles,” likely due to the heavy blur on the sign. Likewise, in the second example, the model predicts “Dog” instead of “Cat”, a plausible mistake given the animal’s appearance. The fourth image is misjudged as a “Bird” rather than an “Airplane,” suggesting that the system recognized a flying object but failed to capture its specific category. This can be attributed to some key features, such as text or outlines, being nearly imperceptible and making the difference between classes difficult to discern. Overall, despite a few misclassifications caused by blurred or partially obscured features, our model successfully distinguishes a wide range of triggers and objects. This highlights its strong robustness against challenging real-world conditions, even when subtle distortions could easily mislead other systems.

\section{Discussion and Limitations}
\label{sec:Discussion}
The empirical results demonstrate that DBOM not only achieves SOA performance in detecting both seen and unseen trigger–object pairings, but also maintains high overall accuracy and attack recall even at low poisoning rates. By proactively vetting training data, DBOM prevents backdoor contamination before downstream model training, reducing the need for costly post‐training purification and preserving clean samples for model learning. 

By separating the backdoor trigger from the underlying object semantics, DBOM not only flags poisoned images, but also recovers the correct object label despite the presence of a backdoor pattern. This has several key benefits. First, it preserves the majority of clean examples so that benign object information is retained rather than discarded, maintaining dataset diversity and reducing the risk of eliminating clean samples. Second, disentanglement yields finer-grained forensic insights into how specific triggers map onto different object categories, revealing systematic attacker strategies and enabling more targeted threat intelligence. Third, the modular nature of trigger and object primitives enables zero-shot detection of trigger-object pairings that were unseen during training, addressing a crucial limitation in conventional trigger-centric defenses. In practice, this means DBOM can adapt to evolving backdoor tactics across multiple object classes, lower false-positive rates by distinguishing benign from malicious features, and streamline training-time vetting helping prevent data contamination at its source rather than reactively purifying a compromised model.

Despite these strengths, it is important to discuss DBOM's limitations. Our design assumes that the defender maintains a library of $T$ candidate trigger patterns, drawn from previously seen backdoor signatures. In our experiments, $T$ is composed of six well-studied backdoor attacks, but the repository can be extended over time as new threats emerge by disentangling unknown triggers and adding them to the trigger repository. When novel trigger patterns are encountered in new data, we can fine‐tune only the visual prompt repository and prefix adapter (rather than retraining the entire VLM backbone) on a small set of those examples, allowing DBOM to rapidly incorporate and detect new triggers with minimal overhead. Although DBOM currently focuses on triggers in $T$, exploring zero‐shot discovery of entirely novel trigger patterns remains an important avenue for future work. Furthermore, the effectiveness of the model depends on the careful tuning of hyperparameters such as $\lambda_{\text{vis}}$, as shown in our ablation study. Moreover, DBOM is currently dependent on VLM encoders, leading to a dependency on the VLM's pre-trained weights. If the VLM fails to classify certain object classes or detect a trigger pattern, then both the visual prompt retrieval and the prefix‐tuned text embedding can be skewed, leading to lower detection rates.  Mitigating this risk in the future may require fine-tuning the VLM on more diverse, trigger‐specific data, or swapping in more powerful multimodal backbones as they become available. However, in this manuscript, we showed base CLIP models are well adept for this task.

While our experiments so far have focused on a select set of backdoor triggers, we have not yet evaluated DBOM against adversarial perturbations generated by methods like Projected Gradient Descent (PGD) \cite{madry2018towards} or Fast Gradient Signed Method (FGSM) \cite{goodfellow2014explaining}. Such attacks work by distributing pixel-level noise within a perturbation budget: when the budget is very small, the changes are imperceptible but often yield lower attack success; when it is larger, the attack becomes more effective but also more noticeable to humans. We believe DBOM’s disentangled trigger–object framework could be extended to handle perturbations with higher budgets, where the noise forms a distinct visual signature similar to the currently tested backdoor patterns and thus can cluster effectively in our visual prompt repository. In future work, we plan to explore these alternative attack types to further test DBOM's resilience. Lastly, evaluating DBOM on larger and more heterogeneous datasets and in real‐world data‐curation pipelines will further validate its practical utility.

\section{Conclusion}
\label{sec:conclusion}
In this paper, we introduced DBOM, a novel disentangled representation learning framework designed to detect both seen and unseen backdoor trigger-object pairings in training datasets. By leveraging a structured factorization of triggers and objects in the embedding space, DBOM enables robust generalization to novel backdoor configurations that evade conventional defenses. Our approach integrates a visual prompt repository and a dynamic prefix adapter to enhance the separation of adversarial triggers from underlying object representations. Experimental results demonstrate that DBOM significantly improves backdoor detection performance, outperforming SOA methods in identifying poisoned samples before they compromise downstream model training. This proactive approach not only enhances the security of DNN training pipelines but also provides deeper insights into backdoor strategies by identifying the objects associated with triggers, offering a novel method for defending against evolving backdoor threats.

\section{Statements} \label{sect:s5}

\acknowledgement{We would like to acknowledge the UWF Argo Cyber Emerging Scholars (ACES) program funded by the National Science Foundation (NSF) CyberCorps® Scholarship for Service (SFS) for making this research possilble.}

\funding{This work is partially supported by the UWF Argo Cyber Emerging Scholars (ACES) program funded by the National Science Foundation (NSF) CyberCorps® Scholarship for Service (SFS) award under grant number 1946442. Any opinions, findings, and conclusions or recommendations expressed in this document are those of the authors and do not necessarily reflect the views of the NSF.}

\authorcontributions{The authors confirm contribution to the paper as follows: 
Conceptualization, Kyle Stein, Andrew A. Mahyari, Guillermo Francia III; 
Methodology, Kyle Stein, Andrew A. Mahyari, Guillermo Francia III; 
Software, Kyle Stein; 
Validation, Kyle Stein; 
Formal analysis, Kyle Stein, Andrew A. Mahyari, Guillermo Francia III; 
Writing—original draft preparation, Kyle Stein, Andrew A. Mahyari, Guillermo Francia III; 
Writing—review and editing, Kyle Stein, Andrew A. Mahyari, Guillermo Francia III, Eman El-Sheikh; 
All authors reviewed the results and approved the final version of the manuscript.}

\availabilityofdataandmaterials{The data that support the findings of this study are available from the Corresponding Author, KS, upon reasonable request.}

\ethicsapproval{Due to the nature of backdoored attacks, all code used to generate them were drawn from methods and datasets already published by the original authors of those backdoor-attack scripts. The backdoored images and their corresponding clean counterparts used in our experiments will be made publicly available after publication to facilitate reproduction of our results. However, we will not redistribute the original attack‐generation scripts. Readers may obtain those directly from the authors of the respective prior works cited in this paper.}

\conflictsofinterest{The authors declare no conflicts of interest to report for the present manuscript.}

\reftitle{References}
\bibliographystyle{vancouver}
\bibliography{ref}  
\end{document}